\documentclass[11pt,a4paper]{article}
\usepackage[hyperref]{emnlp2018}
\usepackage{times}
\usepackage{latexsym}

\usepackage{amsmath}
\usepackage{url}
\usepackage{graphicx}
\usepackage{multirow}

\aclfinalcopy % Uncomment this line for the final submission

\title{Interpretable Textual Neuron Representations for NLP}

\author{Nina Poerner, Benjamin Roth, Hinrich Sch{\"u}tze\\
  Center for Information and Language Processing \\
  LMU Munich, Germany \\
  {\tt poerner@cis.lmu.de}}% \\\And
\date{}

\def\arxiv{1} % set to 1 for arxiv version (attn: re-run bibtex!)

\begin{document}
\maketitle
\begin{abstract}
Input optimization methods, such as Google Deep Dream, create interpretable representations of neurons for computer vision DNNs.
We propose and evaluate ways of transferring this technology to NLP.
Our results suggest that gradient ascent with a gumbel softmax layer produces n-gram representations that outperform naive corpus search in terms of target neuron activation.
The representations highlight differences in syntax awareness between the language and visual models of the Imaginet architecture.
\end{abstract}

\section{Introduction}
Deep Neural Networks (DNNs) have led to advances in Natural Language Processing, but they are hard to interpret.
This is partly due to the fact that their smallest components, i.e., neurons, lack interpretable representations.

For computer vision problems, \citet{simonyan2013deep} propose to use gradient ascent to find an input image that maximizes the activation of a neuron of interest.
Using these image representations, one can for instance show that lower level neurons in vision CNNs specialize in patterns such as stripes \cite{mordvintsev2015deepdream}.

Applying gradient ascent input optimization to NLP is not straightforward, as discrete symbols are not open to continuous manipulation.
A common alternative approach is to search existing corpora for optimal documents or n-grams (e.g., \citet{kadar2017representation}, \citet{aubakirova2016interpreting}).
As this strategy only covers the space of existing inputs, we assume that it may lead to incorrect assumptions.
For instance, the representation of a given neuron may suggest that syntax was learned, when in reality this is due to a lack of ungrammatical inputs in the corpus.
\if\arxiv1
Also, a neuron might attend to a set of concepts that do not usually appear together (e.g., it may fire in the presence of both food-related and sports-related words).
In this case, a search-based representation may only reveal part of the whole picture.
\fi

In the following, we propose and test methods for gradient ascent input optimization in NLP.
Our quantitative assessment suggests that one method, which is based on the gumbel softmax trick, produces inputs that are more highly activating than corpus search.
By applying this method to the Imaginet architecture, we confirm that a language model pays attention to syntax to some degree, while a visual model looks for key content words and ignores function words.

\section{Input optimization for NLP}
In the following, we denote as $f(\mathbf{E})$ the activation of some neuron of interest when forward-feeding a sequence of embedding vectors $\mathbf{E} = [\mathbf{e}_1 \ldots \mathbf{e}_T]$.

\subsection{Embedding optimization}
One straightforward approach to NLP input optimization is to treat $\mathbf{E}$ like \citet{simonyan2013deep} treat images, i.e., to apply gradient ascent directly to the embedding vectors, while keeping other model parameters constant: $\mathrm{argmax}_{{\mathbf{E}}}\big[f({\mathbf{E}})\big]$.
However, there is no guarantee that the optimal vectors will correspond to the embedding vectors of real words, or even be close to them.
In our experiments, the average cosine proximity to the closest real-word embedding is $0.24$, suggesting that there is a divergence between the training goal (finding embedding vectors) and the real goal (finding a representation made up of real words).

\if\arxiv2
To prevent this divergence, we also tried to penalize low maximum cosine proximity by optimizing $\mathrm{argmax}_{{\mathbf{E}}}\big[\mathrm{relu}(r)^\lambda f({\mathbf{E}}) \big]$, where $\lambda$ is slowly increased during training, $r = \frac{1}{T} \sum_{t=1}^T \mathrm{max}_v \big[\mathrm{cos}(\mathbf{m}_v, \mathbf{{e}}_t) \big]$, and $\mathbf{m}_v$ are fixed embeddings of real words. 
We found that the $\mathrm{max}$ term rendered training instable and led to worse results than free embedding optimization, which is why we do not include it below.
\fi

\subsection{Word optimization}
Note that the embedding operation can be written as $\mathbf{E} = \mathbf{X} \mathbf{M}$, where $\mathbf{X} \in \{0,1\}^{T \times V}$ is a matrix of one-hot vectors and $\mathbf{M}$ is the embedding matrix for all $V$ known words.
If we relax the requirement that $\mathbf{X}$ be one-hot, we can perform gradient ascent directly on $\mathbf{X}$, while keeping $\mathbf{M}$ constant: $\mathrm{argmax}_{{\mathbf{X}}}\big[ f({\mathbf{X}} \mathbf{M}) \big]$.
This approach has the undesirable effect that entries in ${\mathbf{X}}$ can become very large or negative, and therefore unlike the one-hot vectors seen in training.

To enforce positive vectors that sum to one, we can use the softmax function across the vocabulary axis: $\mathrm{argmax}_{{\mathbf{X}}} \big[f(\mathbf{P}^\mathrm{smx} \mathbf{M}) \big]$, where $\mathbf{p}^\mathrm{smx}_t = \mathrm{softmax}({\mathbf{x}}_t)$.
However, this input can still be unlike the inputs seen during training, as the optimal distribution may be smooth.

To remedy this situation, we use the gumbel softmax trick (\citet{jang2017categorical}, \citet{maddison2017concrete}): $\mathrm{argmax}_{{\mathbf{X}}} \big[ f(\mathbf{P}^\mathrm{gbl} \mathbf{M}) \big]$, where
\begin{equation}
\nonumber
\mathbf{p}^\mathrm{gbl}_t = \mathrm{softmax}\big[\tau^{-1} \big(\mathrm{log}(\mathbf{p}^\mathrm{smx}_t) + \mathbf{g}_t\big) \big]
\end{equation}
and $g_{t,v} \sim -\mathrm{log}(-\mathrm{log}(\mathcal{U}(0,1)))$.
\if\arxiv1
The resulting probability distribution has the property that selecting its $\mathrm{argmax}$ is equivalent to sampling from $\mathbf{p}^\mathrm{smx}$.
By slowly annealing $\tau$, we are able to transition from a smooth distribution to one where probability mass is highly concentrated, while at the same time avoiding instabilities caused by hard sampling (c.f., \citet{buckman2018neural}).
\else
By slowly annealing $\tau$, we are able to transition from a uniform probability distribution to a ``spiky'' one where probability mass is concentrated on few words.
\fi

\section{Experiment}
\subsection{Model}
We re-implement the Imaginet architecture from \citet{kadar2017representation}.
It consists of a joint word embedding layer (embedding size 1024) and two separate unidirectional GRUs (hidden size 1024 each).
One GRU serves as a language model, while the other predicts visual features of a scene described in the input sentence.
The model is trained on 566435 MSCOCO captions with visual features taken from \citet{chrupala2017representations}\footnote{https://zenodo.org/record/804392/files/data.tgz}.

\if\arxiv0
\begin{figure}[!h]
\centering
\includegraphics[scale=1]{figure_proj}
\caption{Activation after input optimization for randomly selected projection layer neurons. crp: corpus search; emb: embedding opt.; logit: word opt. w/o softmax; smx: word opt. w/ softmax; gbl: word opt. w/ gumbel softmax.}
\label{fig:results}
\end{figure}
\else
\begin{figure}
\centering
\includegraphics[scale=1]{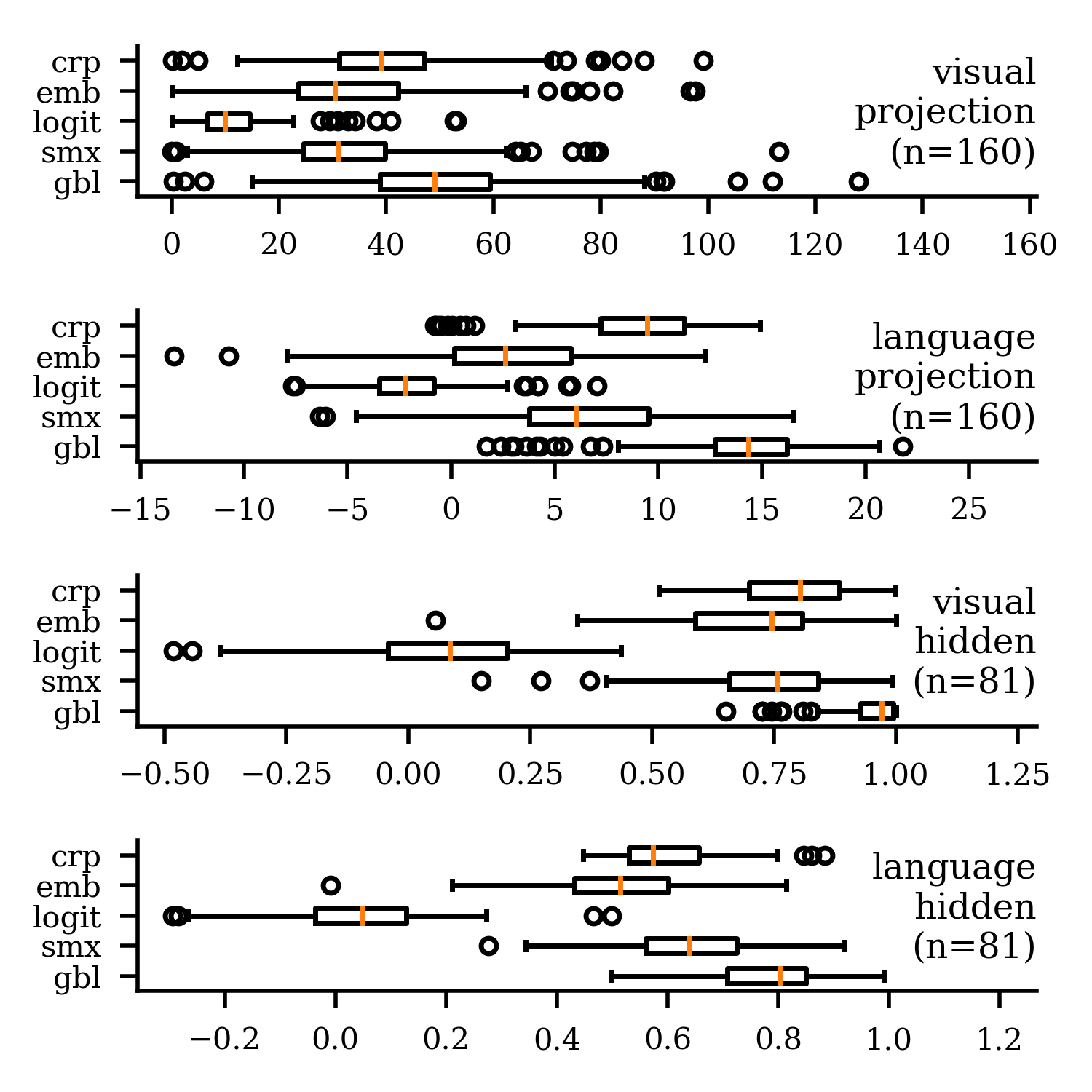}
\caption{Activation after input optimization. crp: corpus search; emb: embedding optimization; logit: word optimization w/o softmax; smx: word optimization w/ softmax; gbl: word optimization w/ gumbel softmax.}
\label{fig:results}
\end{figure}
\fi

\subsection{Quantitative evaluation}
\if\arxiv1
We evaluate the above-mentioned methods by the activation that their optimal representations achieve in target neurons.
We assume that the higher the activation, the better the representation.
\else
We evaluate the input optimization methods by the activation that they achieve in 160 randomly chosen neurons of the language and visual model projection layers.
\fi
For embedding optimization, representations are derived by finding the nearest real-word neighbor of the optimized embeddings in the embedding space. 
For word optimization, we take the $\mathrm{argmax}$ over the vocabulary dimension of ${\mathbf{X}}$.

\if\arxiv1
\subsubsection{Projection layer}
In the projection layer, we randomly select 160 target neurons and find an optimal representation for each one of them individually (Figure \ref{fig:results}, upper boxplots).
Note that in the language model, we maximize the linear pre-softmax score.

\subsubsection{GRU hidden layer}
In the GRU hidden layer, optimizing a single neuron is not very challenging, as the $\tanh$ activation function is easily saturated.
\citet{tangvisualization} report that, contrary to LSTMs, GRUs use highly distributed activation patterns to convey meaningful signals.
Therefore, we evaluate the methods by their ability to achieve high mean activation in disjoint groups of GRU hidden state neurons (Figure \ref{fig:results}, lower boxplots).
The groups are derived by hierarchical clustering with complete linkage.
As distance metric, we use negative activation correlation, as measured on n-grams from the corpus.

\subsubsection{Results}
\fi
We find that while representations from embedding, logit and softmax optimization are not competitive, the gumbel softmax trick outperforms the corpus search strategy in terms of target neuron activation.
\if\arxiv1
Paired t-tests on the difference between corpus search and gumbel softmax representations were highly significant, with $p<0.001$ in all cases: $t=-23.5$ (visual model projection layer), $t=-33.6$ (language model projection layer), $t=-14.1$ (visual model hidden layer), $t=-21.7$ (language model hidden layer).
\fi

\subsection{Qualitative observations}
Table \ref{tab:qual} shows optimal 5-grams for some neurons.

We observe that, contrary to what corpus search suggests, optimal inputs for the visual model rarely contain function words, i.e., the model seems to ignore them.
Optimal inputs for the language model sometimes display grammatically correct structures with function words directly before the predicted word (e.g., ``stare to their [left]'', ``under an [umbrella]'',  see Table \ref{tab:qual}).
This suggests that the language model pays attention to function words and has indeed learned some syntax, as suggested by \citet{kadar2017representation}.

\if\arxiv1
Furthermore, we observe that one neuron may pay attention to different concepts.
For example, the ``race'' neuron in the language model is activated by both horse and motorbike racing words, as evidenced by the gumbel representation (Example 5 in Table \ref{tab:qual}).
The corpus search representation however only reflects horse racing.
\fi

\begin{table}[!h]
\scriptsize
\centering
\begin{tabular}{llr}
method & \multicolumn{2}{c}{optimal 5-gram \hfill target neuron activation} \\ \hline \hline
crp & pizza a sandwich and appetizers	& 48.44 \\ 
gbl & fangs calzone raspberries sandwhich pizzas & 64.46 \\ \hline \hline
\multicolumn{3}{c}{315th neuron in visual projection layer} \\[2pt] \hline \hline
crp & fighter jet flying in formation & 31.25 \\
gbl & propelleor phrases jetliners treetops flight & 37.82 \\ \hline \hline 
\multicolumn{3}{c}{657th neuron in visual projection layer} \\[2pt] \hline \hline
\if\arxiv1
crp & goalie outfit throwing a ball & 65.00 \\
gbl & footplate goalie pitchers racecourse bat & 75.30 \\ \hline \hline
\multicolumn{3}{c}{1006th neuron in visual projection layer} \\[2pt] \hline \hline
\fi
crp & a woman sitting under an & 13.28 \\
gbl & campbell lawn raincoat under an	& 17.54 \\
\hline \hline
\multicolumn{3}{c}{314th neuron (``umbrella'') in language model projection layer} \\[2pt]
\hline \hline
\if\arxiv1
crp & finish line at a horse & 9.45 \\
gbl & horsed horseback motocycles enthusiast they& 13.32 \\ \hline \hline
\multicolumn{3}{c}{522nd neuron (``race'') in language model projection layer} \\[2pt]
\hline \hline
\fi
crp & the view through a car & 10.42 \\
gbl & logging jeep watch through cracked & 14.87 \\
\hline \hline
\multicolumn{3}{c}{957th neuron (``windshield'') in language model projection layer} \\[2pt] \hline \hline
crp & a giraffe looks to its & 10.64 \\
gbl & fest stares stares to their & 13.22 \\ \hline \hline
\multicolumn{3}{c}{973th neuron (``left'') in language model projection layer} \\[2pt]
\end{tabular}
\caption{Examples of optimal 5-grams via corpus search and via gradient ascent with gumbel softmax. Spelling errors stem from the Imaginet dictionary.}
\label{tab:qual}
\end{table}

\section{Conclusion}
The gumbel softmax trick makes it possible to extend the input optimization method to NLP, and to find interpretable textual neuron representations via gradient ascent.
Our experimental results suggest that this technique exceeds naive search on a large in-domain corpus in terms of target neuron activation.
The representations also show interesting differences in syntax awareness based on target modality in Imaginet.
\if\arxiv1
Our code will be made available on \url{https://github.com/NPoe/input-optimization-nlp}.
\fi

\bibliographystyle{acl_natbib_nourl}
\bibliography{emnlp2018}

\end{document}